\documentclass{ieeeaccess}
\usepackage{cite}
\usepackage{amsmath,amssymb,amsfonts}
\usepackage{algorithmic}
\usepackage{graphicx}
\usepackage{textcomp}
\usepackage{caption}
\usepackage{booktabs}
\usepackage{multirow}
\usepackage{cuted}
\usepackage{hyperref}

\def\BibTeX{{\rm B\kern-.05em{\sc i\kern-.025em b}\kern-.08em
    T\kern-.1667em\lower.7ex\hbox{E}\kern-.125emX}}
\begin{document}
\history{Date of publication xxxx 00, 0000, date of current version xxxx 00, 0000.}
\doi{10.1109/ACCESS.2017.DOI}

\title{RANGER: Retrieval Agent for Navigating Graphs with Semantic Evaluation Rewards}

\author{
\uppercase{Artem Mikhaylov}\authorrefmark{1},
\uppercase{Pavel Konoplev}\authorrefmark{1, 2},
\uppercase{Nikolay Efanov}\authorrefmark{1, 2},
\uppercase{Artyom Sorokin}\authorrefmark{3, 1},
and \uppercase{Evgeny Burnaev}\authorrefmark{1, 3}
}

\address[1]{Skolkovo Institute of Science and Technology (Skoltech), Moscow, Russia}
\address[2]{Moscow Institute of Physics and Technology (MIPT), Moscow, Russia}
\address[3]{AIRI, Moscow, Russia}

\tfootnote{The work was supported by the grant for research centers in the field of AI provided by the
Ministry of Economic Development of the Russian Federation in accordance with the agreement
000000C313925P4F0002 and the agreement with Skoltech №139-10-2025-033}


\headname{{\headerfont\color{black}Preprint}}
\markboth{Preprint}{Preprint}

\corresp{Corresponding author: Artem Mikhaylov (e-mail: a.mikhaylov@skoltech.ru).}

\begin{abstract}
The integration of Large Language Models and Reinforcement Learning offers a promising direction for complex sequential decision-making tasks, such as multi-hop question answering over knowledge graphs. While LLMs provide strong semantic understanding and reasoning capabilities, they often struggle with precise, multi-step navigation in structured environments. Conversely, RL agents excel at learning navigation policies but are hindered by the difficulty of designing semantically meaningful reward functions. This paper introduces a novel hybrid approach in which an LLM serves as a mentor to an RL agent during training by providing dense, semantically informed reward signals. The RL agent, built on a Transformer-based policy and trained with Group Relative Policy Optimization, learns to navigate a knowledge graph to retrieve relevant information. The LLM evaluates the semantic relevance of the agent's traversed paths, generating reward signals that align the agent's behavior with the information needs expressed in a query. We integrate this Retrieval Agent for Navigating Graphs with Semantic Evaluation Rewards (RANGER) into a complete question-answering pipeline and evaluate it on the open question answering datasets. Experimental results show that the proposed approach outperforms both a baseline retrieval-augmented system with classical graph search and one using an RL agent trained with heuristic rewards, achieving a significant increase in NLP metrics and LLM-as-a-Judge scores. Our findings demonstrate that deep integration of LLMs into the RL training loop yields more effective and semantically aligned retrieval policies for complex reasoning tasks.
\end{abstract}


\titlepgskip=-25pt

\maketitle

\section{Introduction}
\label{sec:intro}


Recent progress in Large Language Models (LLMs) has precipitated a substantial increase in their functional capacities, facilitating their deployment in complex task execution and their role as the core reasoning engines for autonomous agents operating \cite{wang2024surveylargelanguagemodel}. Despite their remarkable semantic understanding and planning abilities, LLM-based agents often struggle with tasks that require precise multi-step navigation within structured complex environments \cite{r1,r3}, especially in Knowledge Graphs (KGs), where the combinatorial explosion of possible paths and the need for long-term, goal-oriented exploration pose significant challenges for purely LLM-driven approaches, which are often not optimized for sequential decision-making in such discrete, dynamic spaces. Furthermore, they can be computationally expensive and prone to hallucinations when tasked with exhaustive exploration \cite{r1,r3}.

\begin{figure*}[!ht]
    \centering
    \includegraphics[width=0.8\linewidth]{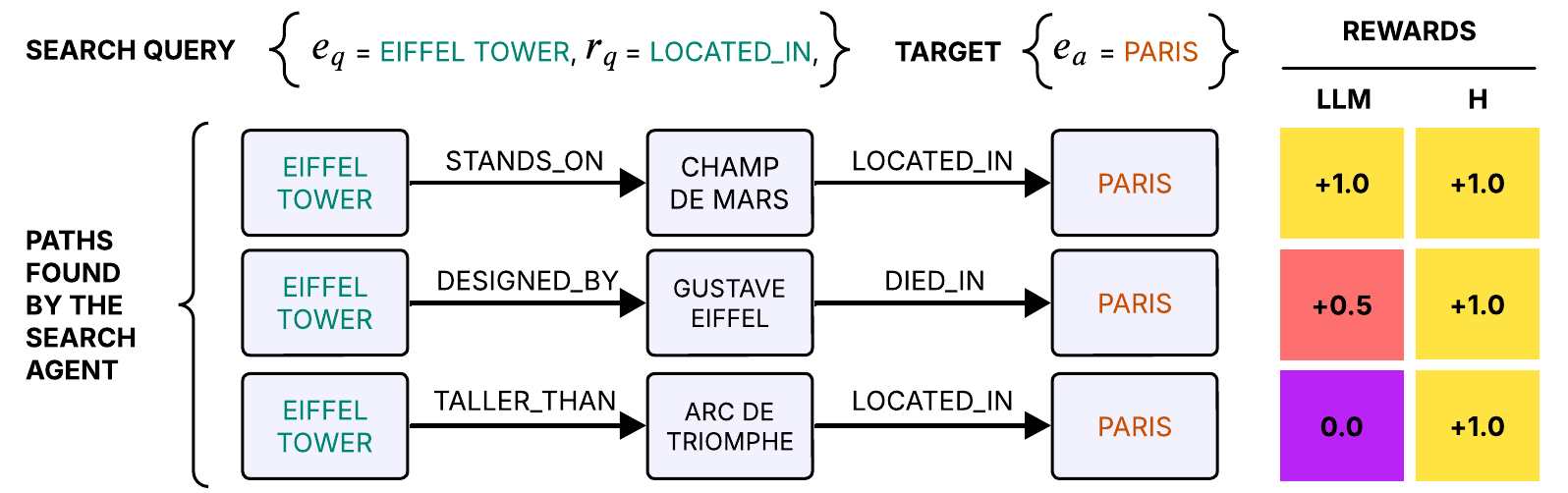}
    \caption{
    Comparison of heuristic and LLM-based rewards for paths in knowledge graphs.
    In this example the question is formed from the triplet 
    $(e_q=\texttt{Eiffel Tower}, r_q=\texttt{located\_in}, e_a=\texttt{Paris})$.
    Given $(e_q, r_q)$, the agent must restore the initial triplet (direct $r_q$ transition is masked out in the KG). The procedural heuristic (\textbf{H}) assigns the same reward to every path that reaches $e_a$, regardless of whether the intermediate triples provide evidence for the original question.
    In contrast, the LLM-based reward (\textbf{LLM}) evaluates whether the discovered path supports reconstructing the original triplet: paths that justify the triplet receive a high reward, while indirect or irrelevant paths receive lower rewards.
    }
    \label{fig:kg_reward_comparison}
\end{figure*}

However, Reinforcement Learning (RL) provides a powerful paradigm for learning optimal decision-making policies through interaction with an environment. RL agents show promising results in tasks involving sequential information search \cite{qrag_2026, rag_rl_2025} and can, in principle, learn to traverse a KG to find relevant information \cite{das2017go, ma2025knowledge}. However, traditional RL faces its own set of formidable challenges: difficulty in credit assignment over long trajectories, and the laborious, often heuristic-driven process of reward function design \cite{r1,r3}. Crafting a reward function that captures not only structural correctness but also the semantic relevance of a discovered path to a user's query is particularly complex.

Recognizing the complementary strengths of these two approaches, this paper explores a novel hybrid architecture that synergistically integrates LLMs and RL. We posit that an LLM can act as a powerful ``mentor'' to an RL agent, guiding its learning process and providing semantic feedback that transcends simple structural heuristics. This integration aims to overcome the limitations of each method in isolation: the LLM provides high-level semantic understanding and a dense, meaningful reward signal, while the RL agent learns a robust and adaptive policy for navigating the structured environment.

In this work, we focus on the challenging task of multi-hop question answering over knowledge graphs. We present a Retrieval Agent for Navigating Graphs with Semantic Evaluation Rewards (RANGER) in which an RL agent, based on transformer architecture and trained with Group Relative Policy Optimization (GRPO), learns to navigate a KG to retrieve information relevant to a user's query.

The main research questions addressed are whether a learned traversal policy can improve the quality of retrieved information compared to classical rule-based search procedures, and whether reward signals produced by an LLM can guide the agent more effectively than rewards based on predefined structural criteria. We also examine how capable the mentor LLM needs to be in order for its feedback to provide an advantage over heuristic rewards.

The key contributions of our work are as follows:
\begin{enumerate}
    \item LLM-as-a-Mentor for Reward Shaping: we move beyond simple heuristic-based rewards and employ an LLM to dynamically evaluate the semantic relevance of the path discovered by the RL agent. This LLM-generated reward provides a context-aware learning signal, directly aligning the agent's navigation policy with the information need expressed in the query. The principle behind this distinction is illustrated in Figure~\ref{fig:kg_reward_comparison}.

    \item Self-Supervised Training from the Knowledge Graph: we demonstrate that effective training of the RL agent can be achieved by automatically generating question-answer pairs directly from the knowledge graph itself. This self-supervised approach reduces reliance on manually annotated datasets and yields a policy that can be used in QA systems for answering complex compositional questions.

    \item Integrated Retrieval Pipeline: we detail the design and implementation of a complete pipeline that integrates our LLM-mentored RL agent as a knowledge retriever within a broader question-answering system. The agent's learned policy guides the extraction of relevant triplets from the KG, which are then used by a separate LLM to generate the final answer.

    \item Empirical Evaluation and Ablation Studies: we compare RANGER with non-parameteric graph traversal algorithms and conduct ablation studies to evaluate the effect of the reward design on downstream QA performance. In particular, we compare the proposed LLM-based reward with a procedural heuristic reward often used in prior KG traversal methods. Our results demonstrate a clear improvement, with RANGER achieving the highest scores across the majority of metrics, including a significant improvement in the LLM-as-a-Judge score.

\end{enumerate}


Our findings underscore the potential of hybrid AI systems in which LLMs and RL agents are not merely combined, but are deeply integrated in a synergistic fashion. Using the semantic prowess of LLMs to shape the learning objective of an RL agent, we can create more effective, adaptable, and intelligent systems for complex information retrieval and reasoning tasks.

\section{Related Work}

Traditional reinforcement learning approaches for knowledge graph navigation laid the groundwork for graph reasoning by formulating it as a sequential decision-making problem on graph structures \cite{das2017go,peng2024graphrag, ma2025knowledge}. The subsequent advent of large language models (LLMs) introduced strong natural language understanding, in-context learning, and reasoning abilities~\cite{r1,r3}, driving their increasing adoption in Knowledge Graph Question Answering (KGQA).
Yet pure LLM approaches suffer from fundamental shortcomings: they lack explicit structured traversal mechanisms, often fail at long-term planning, and tend to generate ungrounded facts~\cite{agrawal2024can, ibrahim2024survey}.

Consequently, LLM-enhanced reinforcement learning has emerged as a promising paradigm to address the inherent limitations of each approach when applied in isolation \cite{r1,r3}, particularly for complex tasks like multi-hop reasoning on knowledge graphs (KGs). This section systematizes the primary methods of integration and situates our work within the context of KG-based Question Answering. 

\subsection{RL-driven Knowledge Graph Navigation}
The initial approaches to using RL for KG reasoning often framed the problem as a classic path-finding task. An RL agent would learn a policy to navigate from a starting entity to a target entity, with the reward function based on heuristic criteria, such as successfully reaching a pre-defined set of answer entities \cite{das2017go}. As noted by \cite{peng2024graphrag, ma2025knowledge}, a significant limitation of these early systems is that their manually designed reward functions, often based on structural graph features, may not capture the semantic relevance of the intermediate steps. The agent might learn to reach a correct entity, but the path taken could be logically unsound or the information gathered insufficient for a complete answer.

\subsection{LLM as a Policy, Trained with Deterministic Objectives}
In the large family of RAG methods, the LLM itself serves as the policy and is trained with reinforcement learning using handcrafted, deterministic rewards. Typical reward functions rely on exact string matching of the answer, word occurrence, or similar lexical heuristics~\cite{r10,r11,r12,r13,r14,r15,r16,r17}. These deterministic rewards are cheap to evaluate and provide an unambiguous optimization signal; however, they are rigid and mostly applicable in training regimes with a single correct ground truth. These metrics cannot recognize paraphrased or semantically equivalent answers and penalize valid reasoning that diverges lexically from the gold path. Moreover, while LLMs serving as the policy are powerful and expressive, they are computationally heavy.

\subsection{LLM as a Mentor}
A rarer paradigm uses an LLM as a mentor or critic that shapes the objective. An LLM-based objective promises a richer, semantically aware learning signal than deterministic heuristics, but it is more computationally expensive, can be prone to hallucinations, and requires careful formulation. The following studies are most similar to our approach.

The authors of \cite{wang2026kg} train an LLM to reason over a KG using an LLM-as-judge reward: after each episode, the LLM evaluates the semantic similarity between the predicted answer and the ground truth. However, the policy learns to trigger a KG retrieval tool rather than to explicitly traverse the graph.
In work \cite{r24}, an actor-critic framework is proposed in which an LLM critic evaluates actions through future rollouts. A key concern is that the policy itself generates the candidate action set, making the framework fragile and prone to reward exploitation. Moreover, the method requires multiple LLM invocations per episode step, incurring heavy computational overhead.
In \cite{zhang2025collaborative}, the authors design a hierarchical framework in which a high-level agent chooses an operation type for a low-level agent that navigates paths. The path reasoning LLM mentor chooses the most promising expansion direction but sees only the history up to the current timestep and immediate actions, restricting guidance to greedy local optima. The approach is also computationally expensive, requiring an LLM call per training step. 
Finally, the work of \cite{wang2025enhancing} uses LLM feedback to motivate the agent to explore more useful information triples. However, the policy requires supervised pretraining on triples that logically support target statements, which are mined using another algorithm. Furthermore, the policy is an encoder-decoder architecture that generates triples in textual form based only on the query triple instead of gathering them from the KG; thus, the approach risks hallucinations and requires supervised retraining whenever the KG changes.

Based on these insights, we introduce a novel reinforcement learning agent based on a lightweight transformer architecture and designed for knowledge graph traversal. It directly navigates the graph, requires no supervised pretraining, and is trained with a single LLM invocation per episode, which incentivizes not merely producing an answer but gathering supporting evidence. Section 4 details its architecture and elaborates on the intricate mechanisms of its integration with LLM, covering the procedures for both training and inference.

\section{Problem Formulation}
\label{sec:problem}

\subsection{Question Answering Task}

We consider a question answering task in a knowledge graph. While questions can be complex and involve multiple dependent conditions, in this section, we focus on a simple type of question with a single clause. Formally, a knowledge graph is defined as a directed multigraph $\mathcal{G} = (\mathcal{E}, \mathcal{R}, E)$, where $\mathcal{E}$ is a set of entities represented as nodes, $\mathcal{R}$ is a set of binary relations between entities, and $E \subset \mathcal{E} \times \mathcal{R} \times \mathcal{E} $ is a set of edges. Then a simple question can be formulated as predicting a destination entity $e_a$ that forms a correct edge $(e_q, r_q, e_a)$, given a query entity $e_q$ and a relation of interest $r_q$. For each query, there may be multiple possible correct answers, for example \texttt{(apple; is a; fruit)} and \texttt{(apple; is a; food)}. In this case we consider finding at least one of them to be a sufficient condition for success. 

To solve the problem, we build a learnable graph traversal algorithm, where the agent sequentially makes transitions between the graph vertices and constructs a continuous path $(e_1, r_1, \ldots, e_n)$ that is decomposed into triplets $(e_i, r_i, e_{i+1})$ constituting the retrieved information. In many works \cite{das2017go, ma2025knowledge} the question answering task is equated to finding a path starting at node $e_q$ and finishing at $e_a$. However, far from every such path in a knowledge graph implies the statement $(e_q, r_q, e_a)$. For instance, a query \texttt{(S. Ramanujan; academic advisor is;  G. H. Hardy)} can not be inferred from a sequence of transitions \texttt{(S. Ramanujan; is known for; Hardy-Ramanujan asymptotic formula; studied by; G. H. Hardy)}. Thus, we argue that in this case, an RL policy often receives positive reinforcement when it should not, and such a formulation is poor for learning information retrieval and reasoning. Besides, in sparse graphs, a situation may arise where the required continuous path does not exist, yet the needed information is still present. Therefore, we pose the task as a search for information supporting the statement $(e_q, r_q, e_a)$. Formulating a strict mathematical criterion for measuring success rate in this task is impractical. Hence, to evaluate the solution quality, we employ large language models, which we discuss further in Section~\ref{Methods}.

\subsection{RL Notation}

 The graph traversal process can be interpreted as a partially observed deterministic Markov Decision Process (MDP) $(\mathcal{S}, \mathcal{A}, \mathcal{O}, \mathbf{T}, \mathcal{R})$, and the agent can be trained using a reinforcement learning algorithm. Below we elaborate on each element of the MDP. 

\subsubsection{States and observations.}

The state $s_t \in \mathcal{S}$ at time $t$ is a tuple $(s^q_t, s^p_t, s^a_t)$, where $s^q_t = (e_q, r_q)$ is a query, $s^p_t = (e_1, r_1, \ldots, e_t),  \; (e_i, r_i, e_{i+1}) \in  E $ is a current path sequence, and $s^a_t = \{e \; | \; (e_q, r_q, e) \in  E \}$ is a set of possible answers to the query. The agent knows the state except for the answers, so its observation is $o_t = (s^q_t, s^p_t) \in \mathcal{O}$.

\subsubsection{Actions.}

The possible actions $A_t \subseteq \mathcal{A}$ at time $t$ are the set of all outgoing edges and incoming edges with reversed relations from the current node $e_t$: $\; A_t = \{ (r, e) \;|\; (e_t, r, e) \in  E \} \cup \{ (r^{-1}, e) \;|\; (e, r, e_t) \in  E \}$. Additionally, we introduce a small set of virtual edges that allow the agent to jump to semantically similar entities, enabling navigation between disconnected components of the graph. Finally, a special ``END'' action is added that terminates the episode.

\subsubsection{Transitions.} 

The transition function $\mathbf{T}: \mathcal{S} \times \mathcal{A} \rightarrow \mathcal{S}$ is deterministic and fully determined by the chosen edge. Specifically, in case of a non-terminal transition, given state $s_t = (s^q_t, s^p_t, s^a_t)$ and action $a_t = (r, e) \in A_t$, the next state is $s_{t+1} = (s^q_{t+1}, s^p_{t+1}, s^a_{t+1})$, where $s^q_{t+1} = s^q_t, \; s^a_{t+1} = s^a_t$ (the query and the set of possible answers remain unchanged), and $s^p_{t+1} = (e_1, r_1, \ldots, e_t, r, e)$ (the path is extended with the chosen relation and entity). If the agent selects the ``END'' action or number of steps exceeds a predefined limit $T$, the episode terminates and the final state is reached.

\subsubsection{Rewards.}

The reward function $\mathcal{R}: \mathcal{S} \rightarrow \mathbb{R}$ assigns a scalar value to each state, incentivizing correct actions and thereby defining the agent's objective. We compare two distinct reward formulations: our proposed LLM-based approach and a heuristic baseline, both described in Section~\ref{Methods}.

\section{Methods}\label{Methods}
\label{sec:methods}

\subsection{LLM-based Reward}

Our proposed method utilizes an LLM to estimate the relevance of the agent's trajectory. It answers the question: ``\textit{Can any of the statements $(e_q, r_q, e_a) \in E$ be logically derived from retrieved information?}''. To enable finer-grained credit assignment, we prompt the LLM to select from three discrete outcomes: ``Success'' (full logical entailment), ``Highly Relevant'' (retrieved information is pertinent but does not directly support the target statements), and ``Failure'' (not even close). These categories are mapped to scalar rewards $r \in \{1, 0.5, 0\}$, respectively. For example see Figure \ref{fig:kg_reward_comparison}.

The main advantage of an LLM as a critic over traditional symbolic or embedding-based metrics is that the LLM can recognize valid logical inferences even when the reasoning chain involves paraphrasing, implicit knowledge, or multi-hop synthesis that does not map cleanly to graph edges. This naturally makes it robust and more applicable to real-world knowledge graphs. Furthermore, it is better aligned with human preferences and captures nuances when evaluating the relevance and coherence of supporting evidence that purely structural similarity metrics might overlook. 

However, one of the notable challenges this approach introduces is high computational cost. For this reason, in this study, the agent receives a non-zero reward value only in terminal states. Another concern is that LLMs can hallucinate, and it is reasonable to expect that smaller language models will yield lower-quality responses. However, as demonstrated in Section~\ref{sec:results}, even a small-sized LLM mentor can match or exceed procedural rewards in terms of downstream utility.

For comparison, we additionally train the RL agent with a heuristic reward function  following prior work on knowledge graph reasoning \cite{das2017go, ma2025knowledge}. Under this formulation, the agent receives a sparse terminal reward $r_t=1$  if the final entity $e_t$  belongs to the answer set $s_a$, and $0$  otherwise. In our experiments, the agent struggles to converge with this sparse reward alone; we therefore help it with a small dense term $\; \min_{a, e_k \in s^p_{t-1}} \operatorname{dist}(e_k, e_a) - \min_{a} \operatorname{dist}(e_t, e_a) \;$ scaled by $w_{\text{dense}} = 0.01$. This term provides a learning signal early in training while having negligible impact near convergence. For RANGER with LLM-based reward, however, this dense component provided no benefit, likely because it incentivizes rapid progression toward answer entities, whereas the LLM reward encourages broader graph exploration. 


\subsection{Policy Architecture}

\begin{figure*}[t]
  \center{\includegraphics[width=0.5\textwidth]{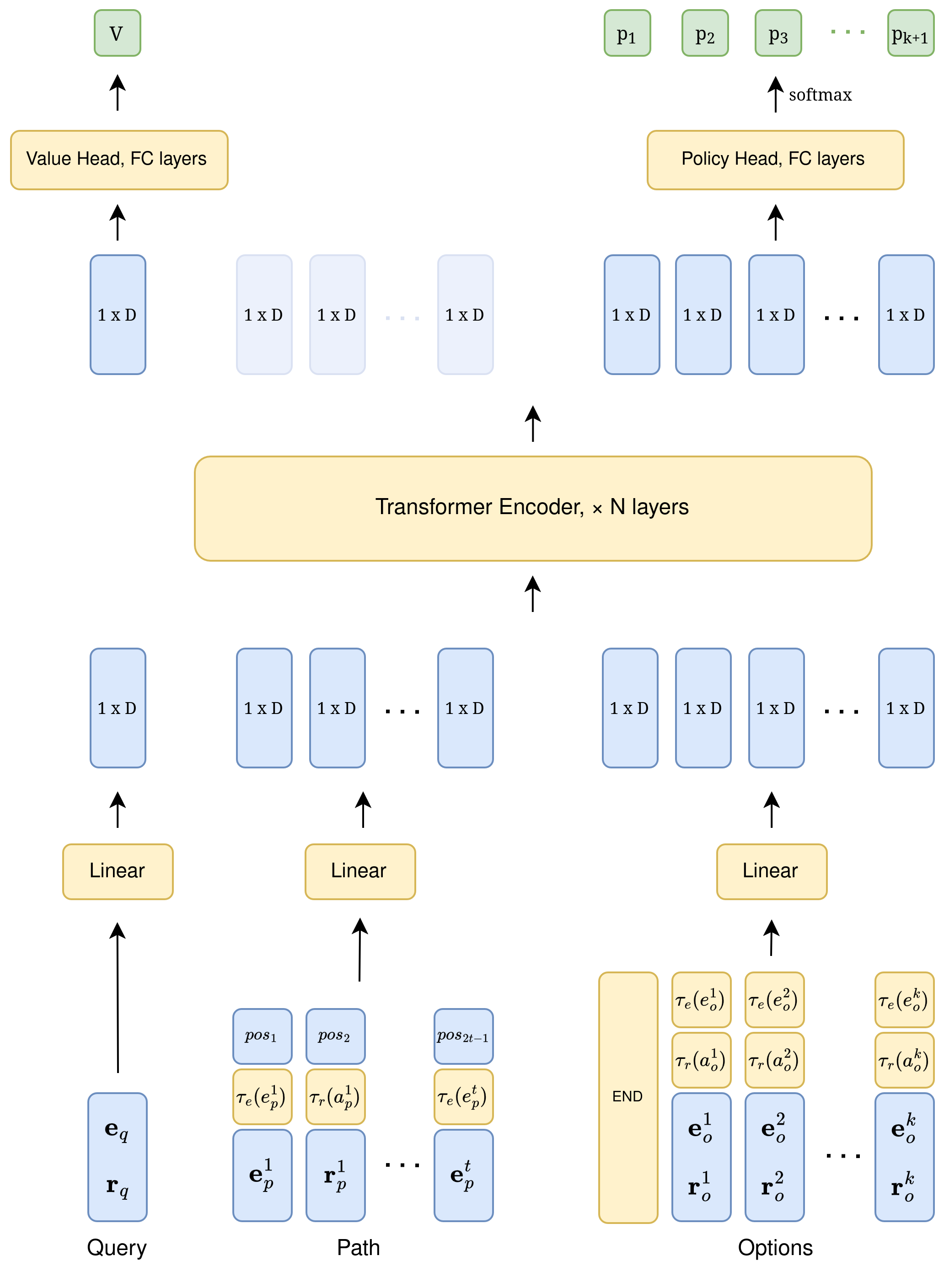}}
  \caption{Agent model architecture. Blocks containing trainable parameters are highlighted in yellow, while output predictions are highlighted in green. The value function head $V$ is used only for PPO algorithm.}
  
  \label{fig:agent_scheme}
\end{figure*}

The policy network in our method is an Actor-Critic model based on the Transformer architecture \cite{vaswani2017attention} and trained from scratch. At each time step $t$, the network receives an observation $o_t = (s^q_t, s^p_t)$ from the environment and the set of available actions $A_t$, and produces a probability distribution $\pi(a_t \mid o_t, A_t)$ over the discrete action space as well as a scalar estimate $V(o_t)$ of the state value. The operating principle of the policy network for a single step is shown in Figure \ref{fig:agent_scheme}.

Textual representations of entities and relations are converted into fixed-dimensional vectors using a pre-trained sentence-transformers embedding model \cite{reimers2019sentence}. The policy network receives three distinct input components:

\begin{enumerate}

\item The query pair $(e_q, r_q)$ is merged into a single vector $[\mathbf{e_q}; \mathbf{r_q}]$, where $\mathbf{e_q}$ and $\mathbf{r_q}$ denote the entity and relation embeddings, respectively, and $[;]$ indicates vector concatenation.

\item Available actions $A_t$ comprise outgoing edges, incoming edges (with reversed relations), and semantic similarity virtual edges. To enable the policy network to distinguish virtual edges from actual ones and identify reversed relations, we introduce learnable action type tokens $ \mathbf{r_{out}}, \mathbf{r_{inv}}, \mathbf{r_{sim}}  \in \mathbb{R}^z$. 
Let $\tau_r: \mathcal{A} \rightarrow \{ \mathbf{r_{out}}, \mathbf{r_{inv}}, \mathbf{r_{sim}} \}$ be the selector function, then for action $a = (r,e)$ we define an augmented relation vector representation as 
$\mathbf{\hat{r}} = [\mathbf{r}; \tau_r(a)]$.

Furthermore, as described below, knowledge graphs in the testing framework contain heterogeneous node types. 
So we also define entity type tokens $\mathbf{e_{obj}}, \mathbf{e_{th}}, \mathbf{e_{ep}} \in \mathbb{R}^z$ and augment entity encodings as $\mathbf{\hat{e}} = [\mathbf{e}; \tau_n(e)]$, where $\tau_n(\cdot)$ is the node type selector function.
The vector representation of an action $a \in A_t$ is then $\mathbf{a} = [\mathbf{\hat{r}}, \mathbf{\hat{e}}]$.
Additionally, a special learnable ``END'' token of the same dimension as other actions is appended to the action set. 

\item Each entity and relation in the current traversal path $s^p_t$ constitutes a separate sequence element. When extending the path after a transition, we keep augmented entity and relation features. Also a sinusoidal positional encoding $pos$ indicating their order in the sequence is mixed-in, yielding the following path encoding:
$\mathbf{P_t} = \bigl( [\mathbf{\hat{e}_1}, pos_1], [\mathbf{\hat{r}_1}, pos_1], \ldots,  [\mathbf{\hat{e}_t}, pos_t] \bigr)$.

\end{enumerate}

The elements of the input sequences are projected to the same dimensionality $\mathbb{R}^{(\cdot) \times d}$ with their own linear transformations:

\[
\mathbf{x_q} = \mathbf{W_q} [\mathbf{e_q}; \mathbf{r_q}], \quad
\mathbf{X_p} = \mathbf{W_p} \mathbf{P_t}, \quad
\mathbf{X_a} = \mathbf{W_a} \mathbf{A_t},
\]

Then all components are combined into a single sequence and processed by a multi-layer Transformer encoder $F_{enc}$:

$$\mathbf{h_q}, \mathbf{H_p}, \mathbf{H_a} = F_{enc}(\mathbf{x_q}, \mathbf{X_p}, \mathbf{X_a}).$$

The self-attention mechanism in $F_{\text{enc}}$ aggregates information from the entire input sequence, yielding global context-aware representations. 
The state value is computed from the query feature vector $\mathbf{h}_q \in \mathbb{R}^d$ via a two-layer MLP:

$$
V_t = \mathbf{w_2}\sigma(\mathbf{W_1} \mathbf{h_q} + \mathbf{b_1}) + b_2,
$$
where $\mathbf{W}_1 \in \mathbb{R}^{d \times d}$, $\mathbf{b}_1 \in \mathbb{R}^{d}$, $\mathbf{w}_2 \in \mathbb{R}^{d}$, and $b_2 \in \mathbb{R}$ are parameters, and $\sigma(\cdot)$ denotes a non-linear activation (ReLU).

Action probabilities are derived from the encoded action representations $\mathbf{H}_a \in \mathbb{R}^{k \times d}$:

$$
\pi_t = \operatorname{softmax} (\mathbf{H_a} \mathbf{W_\pi} + \mathbf{b_\pi}),
$$
where $\mathbf{W}_\pi \in \mathbb{R}^{d}$, and $\mathbf{b_\pi} \in \mathbb{R}^{d}$.  

During training, action choices are sampled from $\pi_t$ to encourage exploration; at inference, we select $\arg\max_{a} \pi_t(a \mid o_t, A_t)$.

\subsection{Extension to General Questions}

The traversal algorithm can, in principle, be easily extended to complex questions involving multiple dependent conditions by encoding the whole question text. However, this would require annotated question-answer pairs.
In this work, we focus on simple questions, as this choice enables unsupervised training of the agent: such queries can be generated procedurally from the knowledge graph itself by sampling existing edges $(e_q, r_q, e_a) \in E$ and masking direct transitions to answers. The resulting algorithm can then be generalized to complex questions by decomposing them into sub-questions, for instance, using a language model, and subsequently aggregating the information retrieved for each simple query to construct the final answer.

\begin{figure*}[t]
  \center{\includegraphics[width=0.8\textwidth]{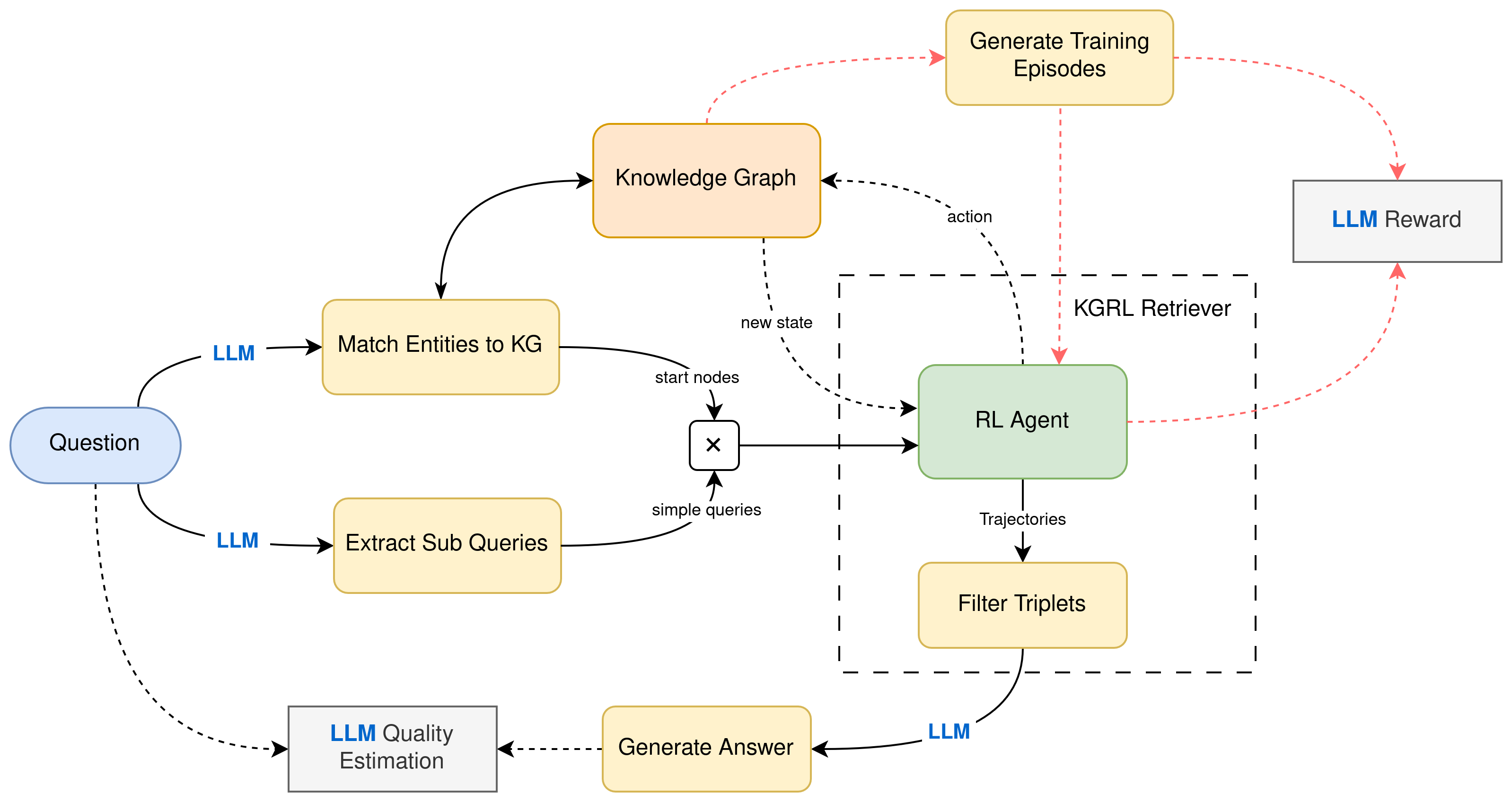}}
  \caption{Answer generation and training pipeline. The agent is embedded into the QA-system retriever component and traverses the knowledge graph to gather information. Red arrows show the training loop. }
  
  \label{fig:pipeline}
\end{figure*}

\subsection{Agent Training Algorithm}

To train the graph traversal policy described above we employ the Group Relative Policy Optimization (GRPO) algorithm \cite{shao2024deepseekmath}. For a given query $q$, the policy $\pi_{\theta_{\text{old}}}$ generates a group of $G$ trajectories $\{\tau_1, \dots, \tau_G\}$. The advantage for the $i$-th trajectory is computed as the normalized deviation from the group mean:

\begin{equation}
\hat{A}_{i,t} = \frac{r_i - \bar{r}}{\sigma_r},
\label{eq:grpo_advantage}
\end{equation}
where $r_i$ is the reward for trajectory $i$, $\bar{r}$ and $\sigma_r$ are the mean reward and standard deviation across the group,  respectively. The policy is then updated using the clipped surrogate objective:

\begin{multline}
L^{\text{GRPO}}(\theta) = \mathbb{E}_{q\sim P(Q), \, \{\tau_i\}_{i=1}^G\sim\pi_{\text{old}}(\cdot\mid q)} \frac{1}{G}\sum_{i=1}^{G}\frac{1}{T_i}\sum_{t=1}^{T_i} \\
\big[\min\big(\rho_{i,t}(\theta)\hat{A}_{i, t}, \operatorname{clip}(\rho_{i,t}(\theta),1-\epsilon,1+\epsilon)\hat{A}_{i, t}\big)\big],
\label{eq:grpo_loss}
\end{multline}
where $\rho_{i,t}(\theta)=\frac{\pi_{\theta}(a_{i,t}\mid o_{i,t})}{\pi_{\theta_{\text{old}}}(a_{i,t}\mid o_{i,t})}$ is the probability ratio for the action taken at timestep $t$ in the $i$-th trajectory for query $q$, and $\epsilon$ controls the clipping range. Unlike the original GRPO formulation, we omit the KL-divergence regularization term, as our policy is trained from scratch rather than fine-tuned from a pretrained reference model.

The choice of GRPO is motivated by the sparse terminal reward signal: since the LLM judge provides feedback only at the end of an episode and, at least early in training, non-zero rewards are rare, training a value function would yield noisy advantage estimates.  For RANGER(H), which combines the sparse terminal reward with the auxiliary dense term, we employ the process-supervised variant of GRPO. Nevertheless, we also conducted an ablation study with Proximal Policy Optimization (PPO) \cite{schulman2017proximal}. Comparative performance of the PPO version is provided in Section \ref{sec:results}.

Given a knowledge graph, we generate training episodes as follows. First, an entity $e_q$ and one of its outgoing edges with relation $r_q$ are selected in a random manner. We then identify the set of all valid answers among outgoing edges where the textual representation of the relation matches $r_q$: $\{e_a \mid (e_q, r, e_a) \in E \, \land \, \operatorname{text}{(r)} = \operatorname{text}{(r_q)}\}$.
The starting node (origin of the path) is chosen in one of two ways. With a predefined probability $p(e_1 = e_q)$, we set the starting node to the query entity $e_q$ itself, excluding direct transitions to answer entities to prevent the task from becoming trivial. Otherwise, we sample a starting node from the neighborhood of $e_q$ at a distance of 2 to 5 hops. Episodes with the first type of starting node can be viewed as knowledge completion tasks, whereas those with the second type correspond to search tasks.

It should be noted that, under edge masking, the knowledge recovery task is not always solvable: if the graph contains no alternative evidence supporting the removed fact, that fact cannot be inferred from the remaining structure. Consequently, the fraction of informative training episodes depends on the completeness and connectivity of the underlying knowledge graph. In our graphs, this fraction appears to be relatively small, as suggested by the learning curves reported in Appendix \ref{app:training_curves}.

\subsection{Integration into Question-Answering Pipeline} \label{PAI_integration}

To test the developed agent in a real-world question-answering scenario, we integrate it into the PersonalAI framework \cite{personalai}. 

The framework is also utilized for building knowledge graphs from contexts. When extracting facts, it creates nodes of three types: object, thesis, and episodic, which differ in the  embedded text volume. The object node most closely corresponds to the conventional concept of an entity, whereas the episodic node contains a brief description. Queries generated for training are allowed to contain only object type entities.

The testing pipeline consists of query preprocessing, knowledge retrieval, and answer generation stages, as illustrated in Figure~\ref{fig:pipeline}.

In the first stage, all entities are extracted from the query using an LLM and are associated with vertices of the knowledge graph via exact matching or similarity in the embedding space. We also added our own operation that extracts a set of simple sub-queries in the form of entity-relation pairs.

The RL agent is incorporated into the retriever component of the framework. It builds $n \times k$ trajectories for each of the $n$ simple sub-queries, starting from each of the $k$ matched entities. Such cross-search can help reveal relationships between entities. The triplets from all trajectories are then collected into a common pool and pruned to a maximum of $N_{max}$ items. For this purpose, we implemented an embedding-based ranking mechanism that assigns high scores to triplets that are semantically similar to at least one of the sub-queries. Finally, the LLM generates a textual answer based on the query and the collected information.

\section{Experiments}
\label{sec:experiments}

The experiments are organized around three aspects. First, we compare standalone RL policies used as retrievers against non-learned graph traversal baselines to gauge their effectiveness for information retrieval. Second, we compare the LLM-based reward with the heuristic reward in terms of answer quality. Finally, we analyze how the size of the LLM mentor used during training affects the quality of the resulting retrieval policy.

As described above, we evaluate the methods within the QA pipeline, as this provides a natural and interpretable way to estimate graph reasoning capability. To isolate the contribution of the retriever, the same query preprocessing, answer generation, and evaluation procedures are used across all runs. For LLM-requiring operations in the generative part of the pipeline, we use Mistral Small 3~\cite{mistral_small31}.

\subsection{Evaluation Metrics}

For each question, the system retrieves evidence from the knowledge graph and generates an answer, which is then compared against the reference answer. We report BLEU-1, BLEU-2, ROUGE-L, METEOR, and Exact Match~\cite{papineni2002bleu,lin2004rouge,denkowski2014meteor} under standard normalization of case and punctuation. To complement lexical-overlap metrics and approximate human review, we perform an LLM-based correctness assessment following the LLM-as-a-Judge protocol~\cite{zheng2023judging}. Given the question, the reference answer, and the generated answer, an independent judge model (Mistral Medium 3~\cite{mistral_medium3}) assigns a score of $1$ for a correct answer, $0$ for an incorrect answer, and $0.5$ for a partially correct answer. The latter category is used, for example, when the answer lists only a subset of the required entities or provides a less specific response, such as a country instead of the expected city.

All metrics are averaged over 150 questions in our experiments.

\subsection{Baselines}

We use two non-parametric traversal algorithms as baselines. The first one is \textbf{Random Walk}, which serves as a lower bound for graph reasoning ability. In this setting, the retriever makes random transitions with the same step budget and starting nodes as the RL agents. The second and more sophisticated method is \textbf{WaterCircles}, the default retriever algorithm used in PersonalAI~\cite{personalai}. It performs multiple breadth-first searches and selects paths with high co-occurrence of the entities mentioned in the question.

\subsection{RL-guided Retrieval}

First, we test RL-guided retrieval on the HotpotQA benchmark, which is designed to evaluate the multi-hop reasoning ability~\cite{yang2018hotpotqa}. For this purpose, we construct a knowledge graph from the HotpotQA contexts and train RL agents on this graph for $10^7$ environment steps in four different configurations. As an ablation study, we consider two reward functions: an LLM-based reward, denoted as \textbf{RANGER(LLM)}, and a heuristic reward, denoted as \textbf{RANGER(H)}. Both variants are trained with two RL algorithms, GRPO and PPO.

We further evaluate a hybrid retrieval setting, motivated by the observation that combining RL-guided traversal with the baseline Watercircles retriever yields substantially greater performance gains than merely increasing the evidence budget of any standalone retriever. In contrast to the standalone scenario, in the hybrid retriever triplets retrieved by WaterCircles are combined with triplets obtained from the RL traversal policy. The resulting triplets are then processed in the standard manner: deduplicated, ranked, and pruned to the same total number of triplets as in all other configurations.

For the hybrid experiments, we consider the same reward functions and RL algorithms as in the standalone setting. To avoid confusion, we use the dagger symbol to denote hybrid configurations: \textbf{RANGER(LLM)$^\dag$} for the LLM-based reward and \textbf{RANGER(H)$^\dag$} for heuristic reward. In addition to HotpotQA, we evaluate the hybrid retriever on TriviaQA and DiaASQ~\cite{joshi2017triviaqa,li2023diaasq}, which represent open-domain factual retrieval and context-dependent dialogue understanding, respectively. The characteristics of the knowledge graphs constructed from these benchmarks are shown in Table~\ref{tab:knowledge_graphs}.

\begin{table}[t]
\centering
\caption{Characteristics of the HotpotQA, TriviaQA, and DiaASQ knowledge graphs.}
\label{tab:knowledge_graphs}
\begin{tabular}{lrr}
\toprule
Dataset & \# Nodes & \# Edges \\
\midrule
HotpotQA & 71,335 & 195,614 \\
DiaASQ & 54,220 & 239,517 \\
TriviaQA & 123,135 & 334,868 \\
\bottomrule
\end{tabular}
\end{table}

Finally, we investigate how many parameters are required in the LLM mentor to provide an advantage over alternative retrieval approaches. The motivation for this experiment is that larger models are generally expected to better capture the context and provide more accurate reward signals, but they also require more computational resources and increase the training time. To study this trade-off, we report the LLM-as-a-Judge score for RANGER(LLM) trained with GRPO using three LLM mentors: Ministral 3 (8B), Mistral Small 3.2 (24B), and Mistral Medium 3.5 (128B). In the experiments above, we use Mistral Small as the default mentor, since, as shown in Section~\ref{sec:results}, it provides a favorable balance between retrieving quality and computational cost.

\begin{table*}[!ht]
\centering
\caption{Comparison of Random Walk, WaterCircles, and RANGER retrieval methods on the first 150 questions from HotpotQA. Best results are shown in bold, and second-best results are underlined.}
\label{table:metrics}

\begingroup
\scriptsize
\setlength{\tabcolsep}{10pt}
\renewcommand{\arraystretch}{1.08}

\begin{tabular}{@{}llcccccc@{}}
\toprule
Method & RL Algorithm & LLM Judge & BLEU-1 & BLEU-2 & ROUGE-L & METEOR & \shortstack{Exact\\Match} \\
\midrule

Random Walk
 & --- & 0.373 & 0.263 & 0.314 & 0.355 & 0.257 & 0.260 \\

WaterCircles
 & --- & 0.673 & \underline{0.455} & \underline{0.324} & \underline{0.578} & \underline{0.452} & \underline{0.467} \\

\addlinespace

RANGER(H)
 & PPO & 0.657 & 0.434 & 0.305 & 0.545 & 0.432 & 0.440 \\

RANGER(LLM)
 & PPO & \underline{0.691} & 0.437 & 0.304 & 0.565 & 0.451 & 0.447 \\

\addlinespace

RANGER(H)
 & GRPO & 0.677 & 0.419 & 0.277 & 0.563 & 0.424 & 0.460 \\

RANGER(LLM)
 & GRPO & \textbf{0.697} & \textbf{0.467} & \textbf{0.340} & \textbf{0.585} & \textbf{0.468} & \textbf{0.493} \\

\bottomrule
\end{tabular}%

\endgroup
\end{table*}

\begin{table*}[!ht]
\centering
\caption{Results of hybrid retrieval methods on the first 150 questions from HotpotQA, TriviaQA, and DiaASQ for GRPO and PPO algorithms. Best results are shown in bold, and second-best results are underlined.}
\label{table:metrics_mixed}

\begingroup
\scriptsize
\setlength{\tabcolsep}{10pt}
\renewcommand{\arraystretch}{1.08}

\begin{tabular}{@{}lllcccccc@{}}
\toprule
Graph & RL Algorithm & \shortstack{Method} & LLM Judge & BLEU-1 & BLEU-2 & ROUGE-L & METEOR & \shortstack{Exact\\Match} \\
\midrule

\multirow{5}{*}{HotpotQA} 
 & --- & WaterCircles & 0.673 & 0.455 & 0.324 & 0.578 & 0.452 & 0.467 \\
\addlinespace
 & \multirow{2}{*}{GRPO} & RANGER(H)$^\dag$ & \underline{0.767} & 0.505 & 0.350 & \underline{0.651} & \underline{0.509} & \underline{0.533} \\
 & & RANGER(LLM)$^\dag$ & \textbf{0.810} & \textbf{0.546} & \textbf{0.367} & \textbf{0.674} & \textbf{0.535} & \textbf{0.540} \\
\addlinespace
 & \multirow{2}{*}{PPO} & RANGER(H)$^\dag$ & 0.743 & 0.507 & 0.344 & 0.630 & 0.498 & 0.520 \\
 & & RANGER(LLM)$^\dag$ & \underline{0.767} & \underline{0.524} & \underline{0.352} & 0.647 & \underline{0.509} & 0.520 \\

\midrule

\multirow{5}{*}{TriviaQA} 
 & --- & WaterCircles & 0.850 & 0.600 & 0.297 & 0.695 & 0.527 & 0.580 \\
\addlinespace
 & \multirow{2}{*}{GRPO} & RANGER(H)$^\dag$ & \underline{0.870} & 0.599 & 0.312 & 0.712 & \underline{0.552} & 0.587 \\
 & & RANGER(LLM)$^\dag$ & \textbf{0.887} & \textbf{0.621} & \textbf{0.322} & \textbf{0.734} & \textbf{0.559} & 0.593 \\
\addlinespace
 & \multirow{2}{*}{PPO} & RANGER(H)$^\dag$ & 0.853 & 0.617 & 0.314 & 0.710 & 0.544 & \underline{0.600} \\
 & & RANGER(LLM)$^\dag$ & \underline{0.870} & \underline{0.619} & \underline{0.321} & \underline{0.721} & \underline{0.552} & \textbf{0.607} \\

\midrule

\multirow{5}{*}{DiaASQ} 
 & --- & WaterCircles & 0.467 & 0.238 & \underline{0.027} & 0.396 & 0.226 & 0.313 \\
\addlinespace
 & \multirow{2}{*}{GRPO} & RANGER(H)$^\dag$ & 0.493 & \underline{0.326} & 0.025 & \underline{0.483} & \underline{0.263} & \textbf{0.447} \\
 & & RANGER(LLM)$^\dag$ & \textbf{0.503} & 0.304 & 0.019 & 0.481 & 0.262 & \underline{0.440} \\
\addlinespace
 & \multirow{2}{*}{PPO} & RANGER(H)$^\dag$ & \underline{0.500} & \textbf{0.333} & \textbf{0.029} & \textbf{0.484} & \textbf{0.267} & \textbf{0.447} \\
 & & RANGER(LLM)$^\dag$ & 0.480 & \underline{0.326} & 0.019 & 0.470 & 0.257 & 0.427 \\
 
\bottomrule
\end{tabular}%

\endgroup
\end{table*}

\section{Results and Discussion}
\label{sec:results}

Table~\ref{table:metrics} presents the QA performance of the baseline and RL-based retrieval methods on the HotpotQA benchmark. Random Walk performs worst across all metrics, showing that unguided graph traversal is insufficient for effective evidence retrieval. WaterCircles is a much stronger baseline and achieves the second-best results for most lexical metrics. However, the learned RL retrievers are able to match or exceed it in terms of answer correctness, which is closely reflected by the LLM-as-a-Judge score. This indicates that training a policy to navigate the graph can improve the retrieval of answer-relevant evidence beyond algorithmic traversal strategies.

RANGER(LLM) trained with GRPO achieves the best results across all reported metrics. It improves the LLM-as-a-Judge score by approximately $3.6\%$ relative to WaterCircles and by approximately $3.0\%$ relative to the GRPO policy trained with the heuristic reward. Among the RL-based approaches, GRPO shows stronger answer-correctness performance than PPO, while the variants trained with the LLM mentor consistently outperform their heuristic-reward counterparts in terms of the LLM-as-a-Judge score. This suggests that the LLM-based reward provides a more effective training signal for learning retrieval policies.

Next, we present the results for the combined retrievers on three benchmarks in Table~\ref{table:metrics_mixed}. In terms of answer correctness, RANGER(LLM)$^\dag$ trained with GRPO achieves the highest on all datasets. Compared with the WaterCircles baseline, it improves this score by approximately $20.4\%$ on HotpotQA, $4.4\%$ on TriviaQA, and $7.7\%$ on DiaASQ. Overall, augmenting the base WaterCircles retriever with RL-guided policies substantially improves the answering ability across all benchmarks. This indicates that the learned policies retrieve complementary evidence missed by the baseline retriever rather than simply duplicating its behavior.

For the hybrid pipelines, the observations regarding LLM guidance and the choice of RL algorithm generally remain consistent with the previous runs: GRPO tends to provide stronger answer-correctness performance, and LLM-based reward variants usually outperform their heuristic-reward counterparts. The main exception to this trend is DiaASQ, where RANGER(H)$^\dag$ under the PPO setting beats other approaches in lexical metrics and is the second-best method in the LLM-as-a-Judge score. This can be attributed to a fundamental difference in reward properties: the heuristic reward is structure-based, whereas the LLM reward is semantics-oriented. Since DiaASQ is constructed from dialogues and contains many semantically similar transitions, evaluating the usefulness of retrieved paths may be more challenging for the LLM mentor.

\begin{figure}[!t]
    \centering
    \includegraphics[width=0.90\linewidth]{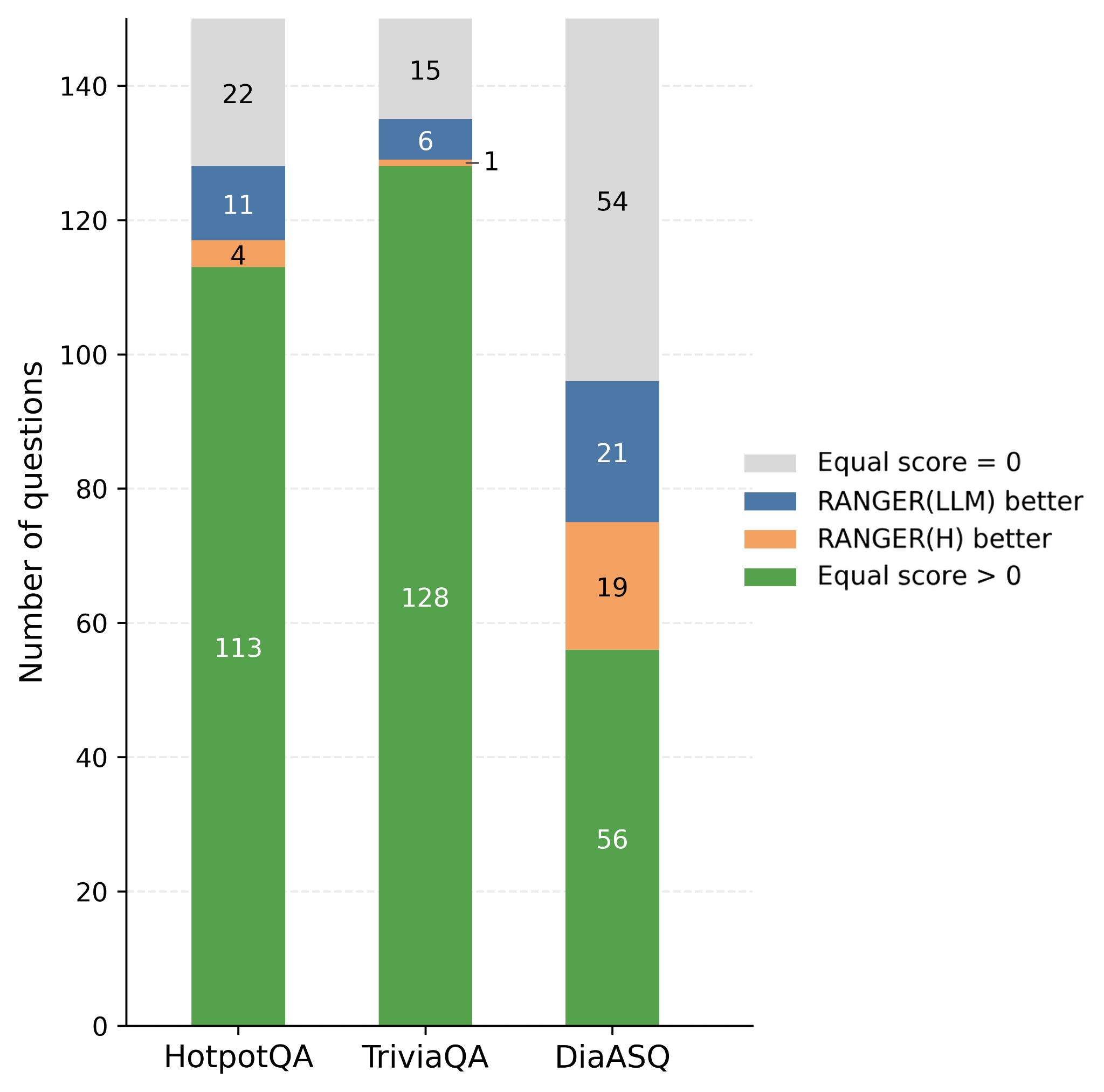}
    \caption{
        Per-question comparison of RANGER(LLM)$^\dag$ and RANGER(H)$^\dag$ using LLM-as-a-Judge scores on the 150 questions from each dataset.
        Each stacked bar partitions questions into four mutually exclusive groups: equal zero scores, RANGER(LLM)$^\dag$ better, RANGER(H)$^\dag$ better, and equal positive scores.
    }
    \label{fig:llm_vs_h_questions}
\end{figure}

Figure~\ref{fig:llm_vs_h_questions} further breaks down the LLM-as-a-Judge results at the question level. On HotpotQA and TriviaQA, the sets of successfully answered questions substantially overlap, as shown by the large number of equal positive scores. However, across all three datasets, and especially on DiaASQ, there are still questions where RANGER(H)$^\dag$ outperforms RANGER(LLM)$^\dag$. This suggests that the LLM-based reward approach has room for improvement, and that a hybrid reward combining semantic feedback with structure-based conditions may lead to further gains.

\begin{figure}[!t]
    \centering
    \includegraphics[width=0.9\linewidth]{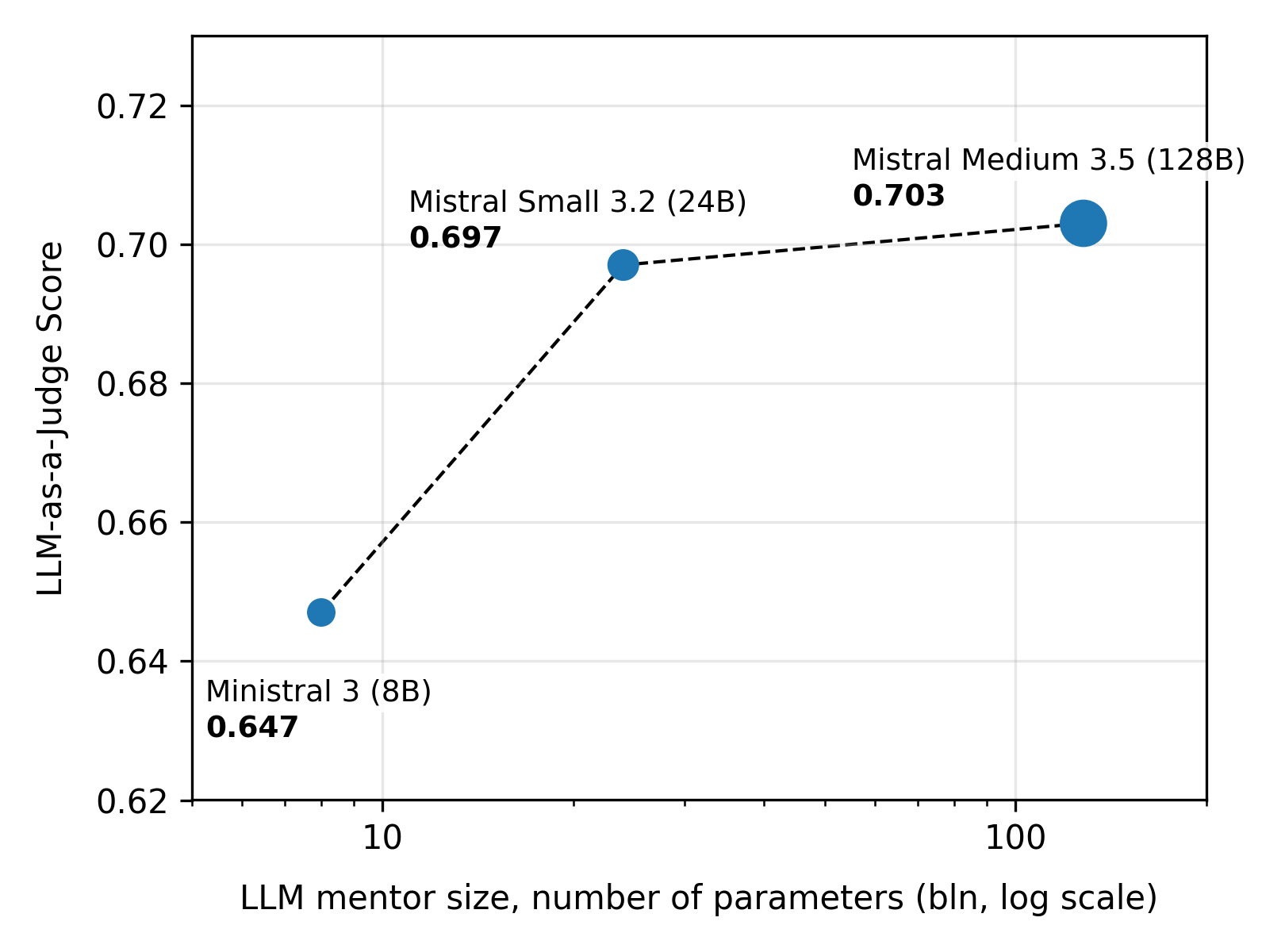}
    \caption{Dependence of the LLM-as-a-Judge score on the size of the LLM Mentor. The score drops substantially from Mistral Small 3.2 (24B) to Ministral 3 (8B), while the improvement from Mistral Small 3.2 (24B) to Mistral Medium 3.5 (128B) is marginal, suggesting diminishing returns from further scaling.}
    \label{fig:score_size_dependence}
\end{figure}

Figure~\ref{fig:score_size_dependence} reports the LLM-as-a-Judge scores for RL policies trained with various LLM mentors. The quality drops substantially when the mentor is reduced from Mistral Small 3.2 (24B) to Ministral 3 (8B), falling below the WaterCircles baseline. However, scaling up to Mistral Medium 3.5 (128B) yields only a marginal improvement over Mistral Small 3.2. Thus, the usefulness of the mentor model increases with model size, but appears to saturate beyond a certain point. The first observation is expected, since the agent partially inherits the reasoning capability of the mentor through the reward signal.

Interestingly, the training curves shown in Appendix~\ref{app:training_curves_llm_mentor} reveal an inverse relation between the mentor size and the magnitude of the assigned rewards. Ministral 3 (8B) gives the most optimistic estimates of retrieved-path relevance, whereas Mistral Medium 3.5 (128B) behaves as the strictest critic. A plausible explanation is that smaller models are more prone to reward-assignment errors, which can make their scores overly optimistic and may also create opportunities for the RL agent to exploit weaknesses in the mentor.

\section{Conclusion}
\label{sec:conclusion}

Based on the experimental results and the presented framework, we conclude that integrating an LLM into the RL training loop through the ``LLM-as-a-Mentor'' paradigm is an effective strategy for multi-hop question answering over knowledge graphs. By using an LLM to provide context-aware reward signals, the approach addresses the limitations of heuristic-based rewards, guiding the RL agent to learn navigation policies that align with the information needs expressed in the query. Importantly, our results also show that the RL agent can be trained in a self-supervised manner directly from the knowledge graph and then used to answer complex questions, without relying on manually curated question-answer datasets. Empirical validation across multiple datasets shows a progressive improvement in both lexical metrics and LLM-as-a-Judge scores, outperforming baseline retrieval-augmented systems and those relying solely on RL with heuristic rewards. These findings suggest that deeply integrated, feedback-driven architectures offer a promising direction for knowledge-intensive reasoning tasks.

\bibliographystyle{IEEEtran}
\bibliography{references}

\clearpage
\begin{appendices}
\renewcommand{\thesubsection}{\thesection.\arabic{subsection}}
\renewcommand{\thesubsectiondis}{\thesubsection.}
\section{Training Details}

\subsection{Learning Curves of RANGER}
\label{app:training_curves}

\begin{figure}[!h]
    \centering
    \includegraphics[width=\linewidth]{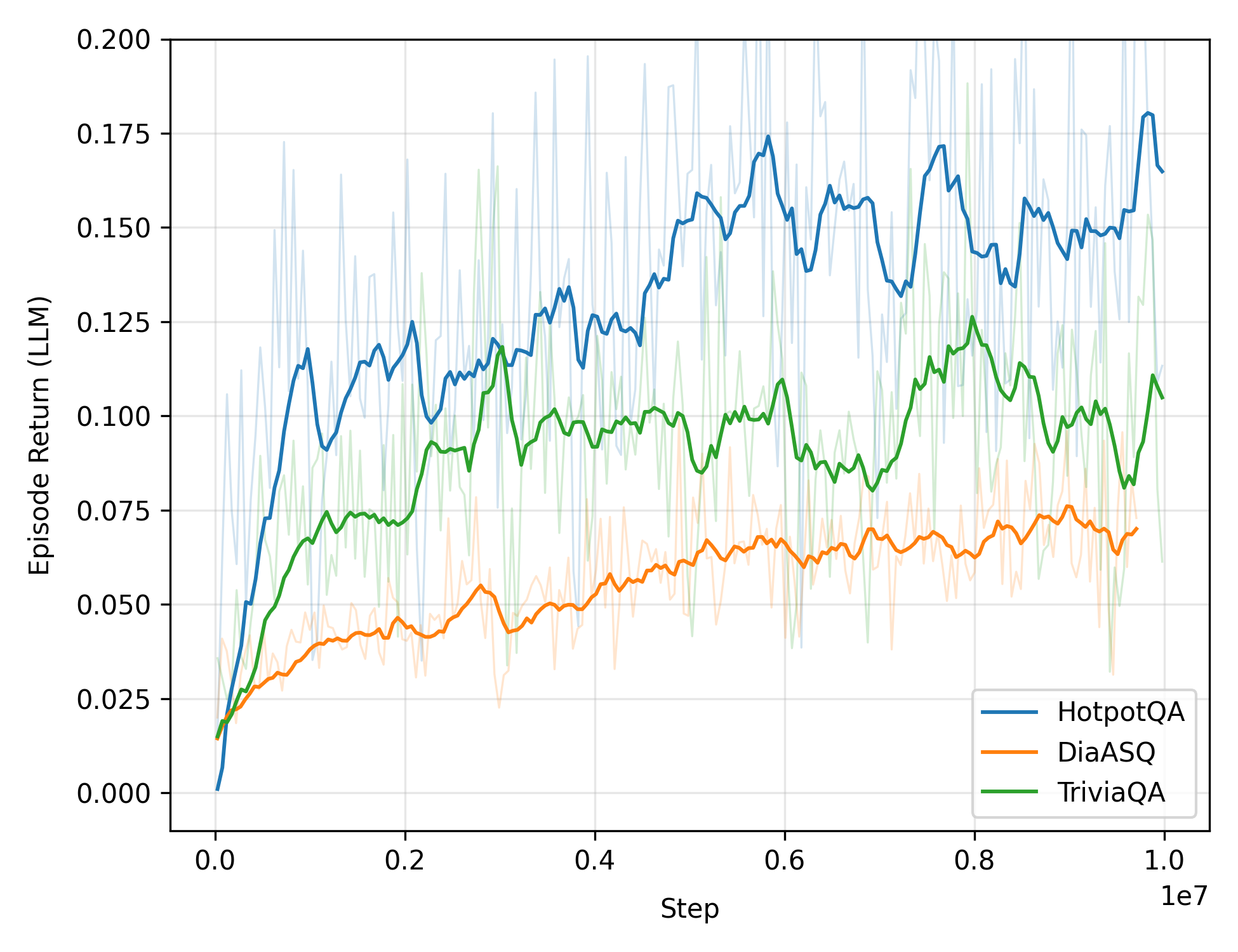}
    \caption{
        Learning curves of RANGER across HotpotQA, DiaASQ, and TriviaQA. 
        The episode return corresponds to the reward assigned by the LLM mentor.
    }
    \label{fig:training-3datasets}
\end{figure}

Figure~\ref{fig:training-3datasets} shows the learning curves of RANGER trained with Mistral Small 3.2 (24B) as the LLM mentor on the HotpotQA, TriviaQA, and DiaASQ graphs.

As shown in the figure, after training for the $10^7$ graph transitions, the average reward assigned to the RL agent by the LLM mentor does not exceed $0.2$ for all graphs. In other words, according to the mentor, the agent finds paths that logically support the target fact in no more than about $20\%$ of episodes. Assuming that the RL agent explores the graph effectively, this suggests that the graphs themselves are sparse in terms of redundant supporting evidence: not every edge is accompanied by alternative paths or evidence that can confirm the corresponding relation.

The learning curves also differ across datasets. In particular, the curve for DiaASQ remains consistently below the others. One possible explanation is that the graphs are constructed automatically, so they may contain noisy, incomplete, or contradictory relations, which can limit the reward achievable by the agent.

\subsection{LLM Mentor Requirements Study}
\label{app:training_curves_llm_mentor}

\begin{figure}[!h]
    \centering
    \includegraphics[width=\linewidth]{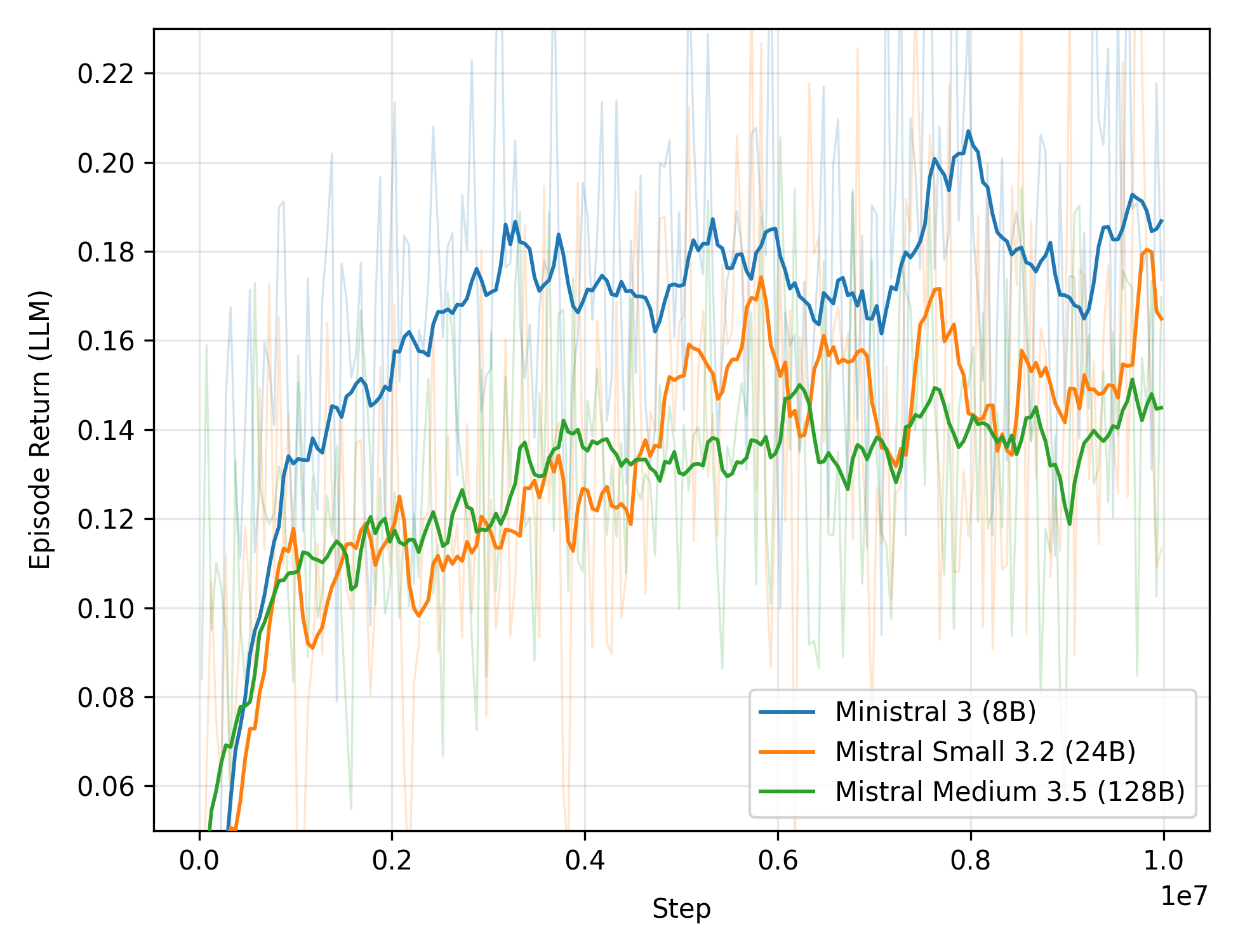}
    \caption{
        Learning curves of RANGER on HotpotQA with Ministral 3 (8B), Mistral Small 3.2 (24B),
        and Mistral Medium 3.5 (128B) as LLM mentors.
        The episode return corresponds to the reward assigned by the LLM mentor.
    }
    \label{fig:training-3models}
\end{figure}

The learning curves of RANGER on HotpotQA with three different mentor LLMs are shown in Figure~\ref{fig:training-3models}.

As can be seen, the average rewards assigned by the mentor during training are highest for Ministral 3 (8B) and lowest for Mistral Medium 3.5 (128B). However, the downstream answer quality shows the opposite trend and increases with the size of the mentor model. This suggests that smaller models may have a weaker understanding of the context and therefore more often assign false positive rewards to retrieved paths.

\section{System Prompt for the LLM Mentor}
\label{app:mentor_prompt}

\begin{figure}[!h]
    \centering
    \includegraphics[width=0.95\linewidth]{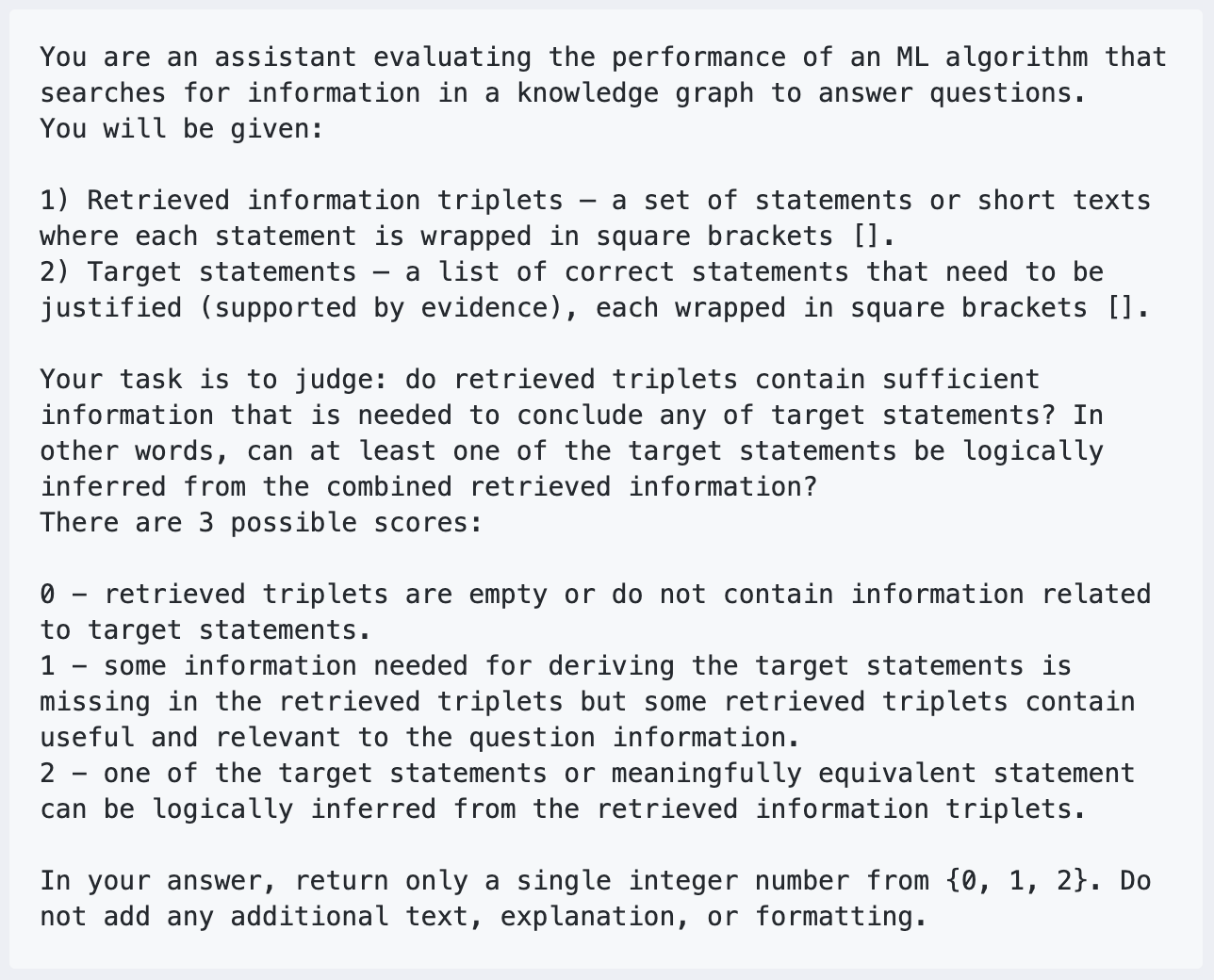}
    \caption{
        System prompt used by the LLM mentor to assign the retrieval reward.
    }
    \label{fig:mentor_prompt}
\end{figure}

The LLM mentor is used to estimate whether the path collected by the RL agent contains sufficient evidence for recovering the target information. The system prompt used for the mentor LLM is shown in Figure~\ref{fig:mentor_prompt}.

\end{appendices}

\eoddefinedtrue 

\end{document}